\documentclass{article}
\usepackage{arxiv}

\usepackage[utf8]{inputenc} % allow utf-8 input
\usepackage[T1]{fontenc}    % use 8-bit T1 fonts
\usepackage{hyperref}       % hyperlinks
\usepackage{url}            % simple URL typesetting
\usepackage{booktabs}       % professional-quality tables
\usepackage{amsfonts}       % blackboard math symbols
\usepackage{nicefrac}       % compact symbols for 1/2, etc.
\usepackage{microtype}      % microtypography
\usepackage{lipsum}
\usepackage{graphicx}
\graphicspath{ {./images/} }

\usepackage[algoruled,resetcount,linesnumbered,ruled,norelsize]{algorithm2e}
\usepackage {algpseudocode}
\usepackage{algorithmicx}
\usepackage{algcompatible}

\usepackage[dvipsnames]{xcolor}
\usepackage{tikz}
\usepackage{pgfplots}
\usepackage{pgfplotstable}
\usepackage{subfigure}
\usepackage{url}
\usepackage{pgf-pie}
\pgfplotsset{compat=1.15}
\usetikzlibrary{patterns}

\title{Object Delineation in Satellite Images}

\author{
 Zhuocheng Shang \\
  Computer Science and Engineering\\
  University of California, Riverside\\
  \texttt{zshan011@ucr.edu} \\
   \And
 Ahmed Eldawy \\
  Computer Science and Engineering\\
  University of California, Riverside \\
  \texttt{eldawy@ucr.edu} \\
}

\begin{document}
\maketitle
\begin{abstract}
Machine learning is being widely applied to analyze satellite data with problems such as classification and feature detection. Unlike traditional image processing algorithms, geospatial applications need to convert the detected objects from a raster form to a geospatial vector form to further analyze it. This gem delivers a simple and light-weight algorithm for delineating the pixels that are marked by ML algorithms to extract geospatial objects from satellite images. The proposed algorithm is exact and users can further apply simplification and approximation based on the application needs.
\end{abstract}

% keywords can be removed

\section{Introduction}
There has been a recent increase in machine learning algorithms and applications that operate on high-resolution satellite data such as land use classification and object detection~\cite{analysis,detection}. This increase has been driven by the public availability of satellite data and the recent advancements in machine learning. Many of these algorithms, such as object detection, produce their output by marking pixels on the satellite data. This output can be enough for regular image processing, such as land field classification. However, it is desirable to delineate pixels corresponding to one object for geospatial applications to form a geospatial polygon that can be further processed in GIS applications. GIS applications generally demand vectorization representations, not individual pixels with geospatial locations. Further, \cite{OpenStreetMap} mentions that machine learning model trained based on OSM data is not accurate due to low quality of vectorial building footprints in the dataset. This drawback motivate the requirement of methods that can delineate pixels into exact rural buildings segments. % update to why openstreetmap need vector representations

Work has been done on a similar challenge that transforms pixels into vector representations~\cite{polygonization}, but the drawback is with approximated boundaries. This gem introduces a light-weight algorithm that delineates marked pixels on a satellite image to produce a valid geospatial polygon, i.e., closed and not self-intersecting. The proposed algorithm is exact in the sense that it exactly delineates all marked pixels with no approximation. Based on its needs, an application can further apply simplification algorithms to produce the desired output. This approach also provides more accurate labeled data in the form of training sets for machine learning models, reducing the misalignment issue~\cite{OpenStreetMap}.

In the graphics field, {\em image tracing} algorithms are widely used to vectorize raster images~\cite{potrace, PolyFit}. These algorithms usually aim to produce basic geometric shapes, e.g., circles and lines, which might entail simplification that is not always desirable for geospatial applications. Vectorized shapes in graphics lack geospatial information, such as latitude and longitute, required in GIS applications. The proposed algorithm is based on the idea of image tracing but is tailored for geospatial data. % change to why graphic filed lack geospatial information

\autoref{fig:overview} gives an overview of the object delineation problem. The gray pixels are the ones marked by the ML algorithm and our goal is to produce the polygon marked by the arrows. To provide an exact answer on satellite data, the output polygons must consist only of orthogonal lines. The key idea is to scan the image once with a $2\times 2$ window to find all the polygon vertices and connect them in the correct order as shown on the figure. The sliding window scans row by row from top-left corner until the bottom right of the canvas. % why 2 x 2 window is enough sufficient

\section{Extracting Objects}
\label{sec:algorithm}

\begin{figure}[htbp]
  \centering
  \includegraphics[width=1\linewidth]{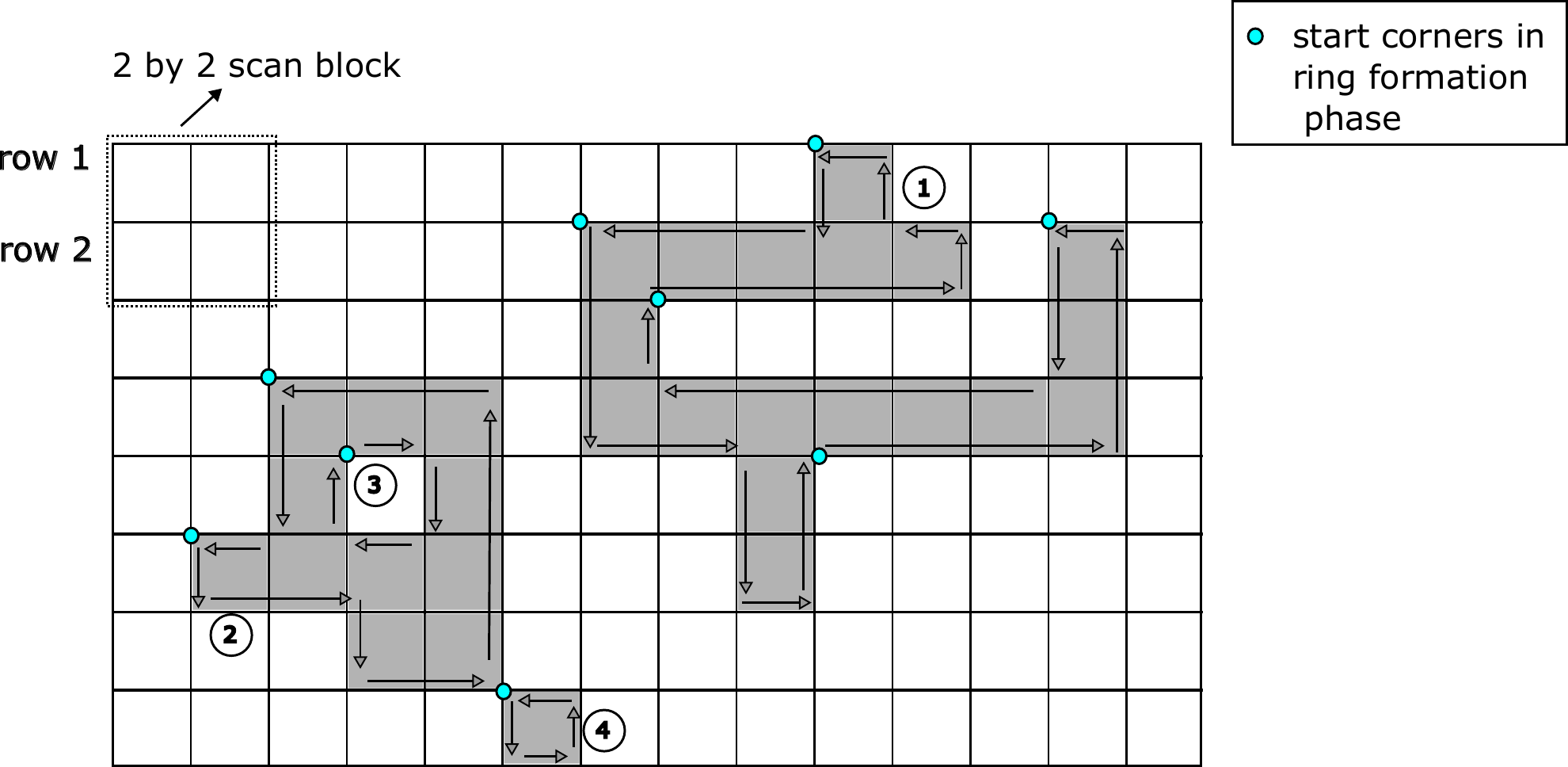}
  \caption{Example of occupied pixels with orthogonal lines }
  \label{fig:overview}
\end{figure}

This section describes two steps in the proposed approach that delineates objects from satellite data, {\em orthogonal lines detection} and {\em ring formation}. The detection step locates all orthogonal lines from occupied pixels in a single scan over the image. The ring formation step combines orthogonal lines into geospatial linear rings.

\autoref{fig:overview} gives an overview of what the algorithm does. Given a raster image with marked pixels, it creates an orthogonal polygon that surrounds all marked pixels. The vertices that make polygons are all located at pixel corners as shown in figure. By convention, the vertices of a polygon is ordered in counter clock-wise order (CCW). In case of a polygon with a hole, the vertices of the hole are ordered in clock-wise order (CW).

\begin{figure}[htbp]
  \centering
  \includegraphics[width=0.85\linewidth]{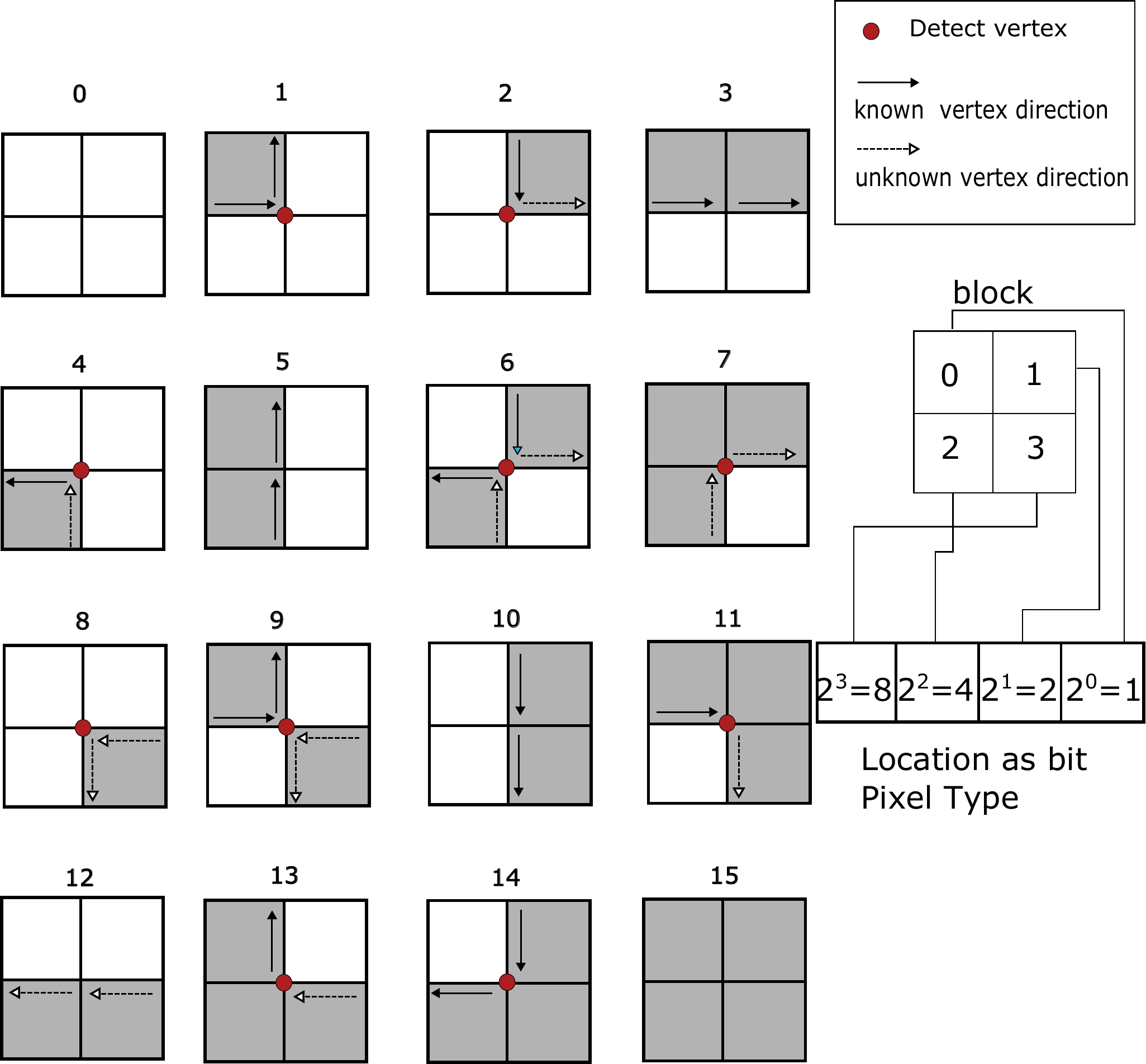}
  \caption{Different pixel occupied cases within one 2 by 2 block}
  \label{fig:cases}
\end{figure}

\subsection{Orthogonal Lines Detection}
\label{sec:detection}

%% create edges for occupied pixels
%% figures of 16 cases
This step takes the marked raster as input and produces all the orthogonal lines that comprise all the polygons. The lines are grouped in rings, i.e., a circular linked list of vertices, as shown in \autoref{fig:overview}. The key observation is that each vertex on the polygon connects a horizontal edge to a vertical edge. Thus, to find all vertices, we need to locate the parts of the image where a horizontal edge meets a vertical edge. Then, we should connect these vertices in the correct order to create a closed polygon. The problem becomes particularly challenging when dealing with many polygons and high-resolution satellite images where a group of few pixels could form one ring. In such a situation, the naive image-tracing algorithm needs to track many small polygons which complicates the algorithm. Therefore, this gem intends to deliver a light-weight algorithm that efficiently finds all polygon segments as orthogonal lines on the high-resolution image through a single scan.

In order to overcome the issue of tracking all orthogonal lines, we observe that a vertex can be detected by checking the $2\times 2$ window where the vertex is at the center. This is sufficient to detect that a horizontal and vertical edges meet. \autoref{fig:overview} illustrates an example $2\times 2$ window. This step simply runs a sliding window over the entire raster to create all the vertices and arrange them in the correct order as further detailed below. The sliding window starts from the top-left and slides over each row from left to right.

%The purpose of this phase is illustrated in \autoref{fig:overview} with two examples shadowed in grey, and the process intends to detect all lines with directed arrows shown on the raster canvas. The implemented algorithm begins by scanning the raster canvas row by row with a scan window arranged 2 by 2 pixels. \autoref{fig:overview} presents more details, and it indicates the scan window starts at the first row highlighted with a dashed square, and the subsequent detection block is next to it. Likewise, row 2 is the start position at the second line. Meanwhile, the approach provides an orthogonal line detection method within each scan window. 

Since a $2\times 2$ window contains only four pixels, and each pixel can either be marked or unmarked, there are a total of 16 possible cases that can happen. If we handle all of them correctly, then we know that we have all possible cases covered. \autoref{fig:cases} illustrates all the 16 cases. For formalization, we identify these cases by assigning a bit position to each of the four pixels as show in the figure. We can immediately see that cases 0, 3, 5, 10, 12, and 15 do not result in any detected vertices. Cases 1, 2, 4, 7, 8, 11, 13, and 14, each result in creating one vertex at the center of the window. Cases 6 and 9 are special cases that result in two coinciding vertices, both at the center. We chose to create these two vertices to ensure that we create non-intersecting and non-overlapping polygons. If our goal is to only find the location of the vertices without caring about their connection and order, then it is enough to scan all $2\times 2$ windows and emit a vertex for each of the above cases. However, we also want to connect them in the correct order which we describe below.

We notice that each vertex must connect a horizontal edge to a vertical edge. The horizontal edge must be at the same row, and the vertical edge must be at the same column. Following our convention, the direction of the created edges is shown in \autoref{fig:cases}. However, we notice that we cannot always create the edge when we create a vertex. For example, Case~1 creates a vertex that connects another vertex to the left to a vertex to the top. This is an easy case because both vertices must have been already created following our sliding window order, i.e., left-to-right and top-to-bottom. However, Case~2 is harder since it connects a vertex to the top to a vertex to the right that is not yet created. Hence, while handling Case~2, we can only create the vertical edge and not the horizontal edge because we do not know the position of its next vertex. In Case~7, we cannot create any edge since the end points of both edges are not known yet. In \autoref{fig:cases}, solid arrows indicate the edges that can be created in each case while dotted arrows indicate edges that cannot be created at that point.

To resolve the issue of incomplete edges, we define the notion of an {\em open vertex}. An open vertex is part of an edge that is not yet created, i.e., the other end point is not detected yet. Since all edges are orthogonal, each open vertex can only be paired with another vertex at the same row or the same column. In addition, given the order at which the $2\times 2$ window slides, a vertical edge can only have an open vertex at the top and a horizontal edge can only have an open vertex to the left. Therefore, as we scan, we keep at most one open vertex to the left and at most $w+1$ open vertices at the top, one for each column, assuming the raster data has a width of $w$ pixels.

Given these open vertices, all edges can be created efficiently in one scan. If a vertex is the top vertex of a vertical edge or a left vertex of a horizontal edge, it is stored as an open vertex. On the other hand, if it is the bottom vertex of a vertical edge or a right vertex of a horizontal edge, it is paired with the corresponding open vertex. During this scanning, each vertex is stored in one circular linked list, and multiple such linked lists could exist on one raster image, as shown in \autoref{fig:overview} which contains four circular linked lists. The following content gives more details about forming all circular linked lists connecting all vertices in the correct order.
%Before instantly explaining detecting details, three concepts are required to be clarified. Firstly, the understanding of the scan window operated. Each cell in one scan block is indexed with a position number, illustrated in \autoref{fig:cases}, and holds a bit value converted from binary representation. Each sum of combinations between bit values represents an individual case illustrated correspondingly with a case number. Therefore, the figure offers all 16 possible combinations of different pixel numbers and their occupied locations in such a 2 by 2 block. Another concept required to be specified is that in this algorithm, we define orthogonal lines as horizontal and vertical lines connected, and the corner they meet as detection vertex $\langle x, y\rangle$ symbolized as red points in \autoref{fig:cases}. Moreover, another assumption is that one occupied pixel is connected to four pixels to its north, west, south, and east. Thus, the algorithm stores all detection vertices into one linked list with $\langle x, y, next\_vertex, visited\_status\rangle$ on each node where the third term presents the next connected pixel and the fourth term is used in the merging phase.

In \autoref{fig:cases}, Case~1 creates a vertex that connects the left vertex to the top vertex. Both left and top vertices must be open vertices that have been created before reaching that case but without knowing the other end point. Thus, we will update the next link of the left open vertex to point to the newly created vertex and update the next link of the new vertex to point to the top vertex. On the other hand, Case~2 connects the top vertex to a vertex to the right that is not reached yet. Therefore, we updated the next link of the vertex at the top to point to the new vertex. Then, we keep this new vertex as an open vertex to the left to be paired later. In Case~7, the newly created vertex is both an open left vertex and an open top vertex so it will be stored as that.

Cases~6 and~9 are more interesting. It indicates two pixels meeting at a corner. To keep the created polygons valid, we create two coinciding vertices at the center. In other words, Case~6 does the work of Cases 2\&4 together and Case~9 combines Cases 1\&8 together.

Algorithm~\ref{alg:detecting} gives the pseudo-code of detecting orthogonal lines. The input is a two-dimensional bit array of width $w$ and height $h$. The output is a set of circular linked lists of vertices. Each vertex has an integer coordinate $(x,y)$, a pointer to the {\em next} vertex, and Boolean {\em visited} flag that will be used in the next algorithm. The output is stored in a list of {\em corners} that contains at least one pointer for each linked list.

We keep a list of open vertices at the top and a single pointer to the left open vertex. All these are initialized to null. We run a loop over all pixels that slides a window at the center of each intersection $(x,y)$ on the raster grid. Then, it computes the pixel type $[0,15]$ by inspecting the four pixels in the window. Any pixel that falls outside the raster grid is assumed to be non-marked. After that, it runs a single switch statement that efficiently handles all the cases. To keep track of all linked lists, we need to store at least one pointer in each circular linked list, and we define this pointer as \emph{start corner}. This preserved \emph{start corner} list is used in the next phase to help with ring formation. We chose to record the top-left corner only which is handled by Cases 7,8, and 9. There is no necessity to record all vertices, because the top-left vertex is enough for pixel location representation. Further, all polygons contain one top-left vertex, and this gives the reason selecting case 8 and 9 is presentable. The necessity of storing case 7 is to correctly handle polygon with holes. For example, if we intend to output the polygon holds a hole in \autoref{fig:overview} labeled 3, we also need to keep the top-left corner for the pixel illustrated as one hole. Therefore, one polygon requires multiple start corners so that ensures formatting boundaries correctly. % add explanation

%Switch cases of $pixelType$ are numbered the same as in \autoref{fig:cases}. The pseudo-cod takes an input of a sequence of pixels as represented with its top-left vertex $\langle x,y\rangle$ and one raster canvas initialized with width and height. The output is an array of linear rings. 

To analyze the time complexity of this algorithm, we note that the cost is mainly in the for loop that iterates over each pixel. The switch statement has a constant-time cost. Thus, the time complexity is $O(w\cdot h)$ which is linear in terms of number of pixels. For space complexity, we observe that, in addition to the input, we need to keep track of open vertices at the top which requires $O(w)$ space. Also, we need to keep track of all the vertices which requires $O(|V|)$. So, the space complexity is $O(|V|+w)$ which is output-sensitive.

%This algorithm computes with a linear running time component of the complexity of final output vertices \emph{V} and the raster canvas size \emph{R}. Hence, the running time is \emph{O(V+R)}. Memory usage is also dependent on the complexity of output vertices \emph{V}.  

\begin{algorithm}
\label{alg:detecting}
\SetKwInOut{Input}{Input}
\SetKwInOut{Output}{Output}
\Input{R: Two-dimensional bit array [$w$][$h$] for marked pixels}
\Output{Linked List of vertices $\langle x: Int, y: Int, next: Vertex, visited=false\rangle$\\ List of start Corners $\langle vertices \rangle$}
    topVertices: Array of open vertices at each column of size $w+1$\\
    leftVertex: The open vertex to the left or null \\
    Corners: List of start corners\\
\For{$y\in [0,h]$, $x\in [0,w]$} {
    block 0 $\gets$ 1 if R[x-1][y-1] not empty, otherwise 0\\
    block 1 $\gets$ 2 if R[x][y-1] not empty, otherwise 0\\
    block 2 $\gets$ 4 if R[x-1][y] not empty, otherwise 0\\
    block 3 $\gets$ 8 if R[x][y] not empty, otherwise 0\\
    pixelType $\gets$ (block 0 + block 1 + block 2 + block 3)\\
    \Switch{pixelType}{
    {\bf case} 0,3,5,10,12,15 {\bf do} nothing\\
    {\bf case} 1 {\bf do} leftVertex.next $\gets$  V$\langle x, y, topVertices(x)\rangle$ \\
    {\bf case} 2 {\bf do} leftVertex $\gets$ topVertices(x).next $\gets$ V$\langle x, y, null\rangle$ \\
    {\bf case} 4 {\bf do} topVertices(x) $\gets$ V$\langle x, y, leftVertex\rangle$\\
    \Case{6}{
    v1 $\gets$ topVertices(x).next $\gets$ V$\langle x, y, null\rangle$ \\
    topVertices(x) $\gets$ V$\langle x, y, leftVertex\rangle$\\
    leftVertex $\gets$ v1}
    {\bf case} 7,8 {\bf do} Corners $\ll$ topVertices(x) $\gets$ leftVertex $\gets$ V$\langle x, y, null\rangle$\\
    \Case{9}{
        leftVertex.next $\gets$ V$\langle x, y, topVertices(x)\rangle $ \\
        Corners $\ll$ topVertices(x) $\gets$ leftVertex $\gets$ V$\langle x, y, null\rangle$\\
    }
    {\bf case} 11 {\bf do} topVertices(x) $\gets$ leftVertex.next $\gets$ V$\langle x, y, null\rangle$ \\
    {\bf case} 13 {\bf do} leftVertex $\gets$  V$\langle x, y, topVertices(x)\rangle$\\
    {\bf case} 14 {\bf do} topVertices(x).next $\gets$ V$\langle x, y, leftVertex\rangle$
    }
}
\caption{Orthogonal Lines Detection}
\end{algorithm}

\subsection{Ring Formation}
\label{sec:merge}

%% combine into rings
This step takes as input the circular linked lists created by the first step and combines each one into a single ring through one start corner list generated in the previous step. Based on how the rings were formed, outer rings and inner holes are ordered in CCW and CW order, respectively. To complete merging in one round, the implemented algorithm activates by picking one start corner from a preserved list, shown as dots in \autoref{fig:overview}, and then goes over the entire list. Each vertex in the list is converted to a geospatial coordinate $(longitude,latitude)$ and is combined to produce the geospatial ring. To convert integer raster coordinates $(x,y)$ to geospatial coordinates $(longitude,latitude)$, we use an affine transformation, termed grid-to-world, and this is a standard method to encode geospatial coordinates of raster datasets \cite{raptor}. While iterating over the vertices, they are marked as visited by setting the flag in each vertex. This ensures that each ring is converted only once since one ring can contain multiple corners, e.g., ring~1 in \autoref{fig:overview}. After one ring is formed, it is appended to a list of rings which are then returned by the algorithm.

%The goal of this work is to provide geospatial polygons, hence, the spatial coordinates are calculated through the conversion method: grid integers $\langle x, y\rangle$ to world format $\langle latitude, longitude\rangle$. Meanwhile, the saved vertex is updated as visited. A geometry linear ring will be created when the next vertex is the same as the start corner, which means the dedicated coordinates form a blocked region. We repeat the step until loop through all the start corners preserved. Algorithm \autoref{alg:merging} pseudo-code shows the general merge process and returns an array of geospatial linear rings.

Algorithm~\ref{alg:merging} provides the pseudo-code of the ring formation process. The input is the list of start corners created by the first step. It loops over all start corners that are not yet visited. For each start corner, it follows the linked list until it goes back to the start since it is a circular linked list. It converts each vertex to geospatial coordinates using the grid-to-world transformation. Finally, it appends the first point again to close the ring and appends it to the list of rings.

To analyze the time complexity of this algorithm, we observe that the major part is going through all vertices stored in all the linked lists. The time complexity is linear in term of number of vertices stored, which is $O(|V|)$. Thus, the time complexity is output-sensitive. Both the input and output sizes are equal to number of vertices so the space complexity is also $O(|V|)$.
% Note: There is no need to use the number of rings since it is always smaller than the number of vertices.

\begin{algorithm}
\label{alg:merging}
\SetKwInOut{Input}{Input}
\SetKwInOut{Output}{Output}
\Input{Corners: List of start Corners\;}
\Output{Rings: List of Geometry Linear Rings}
\SetKwRepeat{Do}{do}{while}
\For{corner $\gets$ Corners}{
  \If{{\bf not} corner.visited}{
      p $\gets$ corner\;
      Coordinates$\gets\langle\rangle$: List of geospatial coordinates\\
      \Do{p != corner}{  
          (longitude,latitude) $\gets$ grid\_to\_world($p.x,p.y$)\\
          Coordinates $\ll$ ($longitude$, $latitude$) \\
          p.visited $\gets$ true \\
          p $\gets$ p.next\\
      }
      
      (longitude,latitude) $\gets$ grid\_to\_world($p.x,p.y$)\\
      Coordinates $\ll$ ($longitude$, $latitude$) \\
      Rings $\ll$ Coordinates
  }
  \Return Rings
}
\caption{Ring Formation}
\end{algorithm}

\section{Experimental Result}
\label{sec:result}

\begin{table}
 \caption{Experimental test cases setup}
  \centering
  \begin{tabular}{ll}
    \cmidrule(r){1-2}
    Raster W$\times$H     & Probability of marked pixel ($p$)  \\
    \midrule
    1000 $\times$ 1000 & [ 0.0 , 1.0 ]  \\
    2000 $\times$ 2000 & [ 0.0 , 1.0 ] \\
    4000 $\times$ 4000 & [ 0.0 , 1.0 ] \\
    
    \bottomrule
  \end{tabular}
  \label{tab:table}
\end{table}

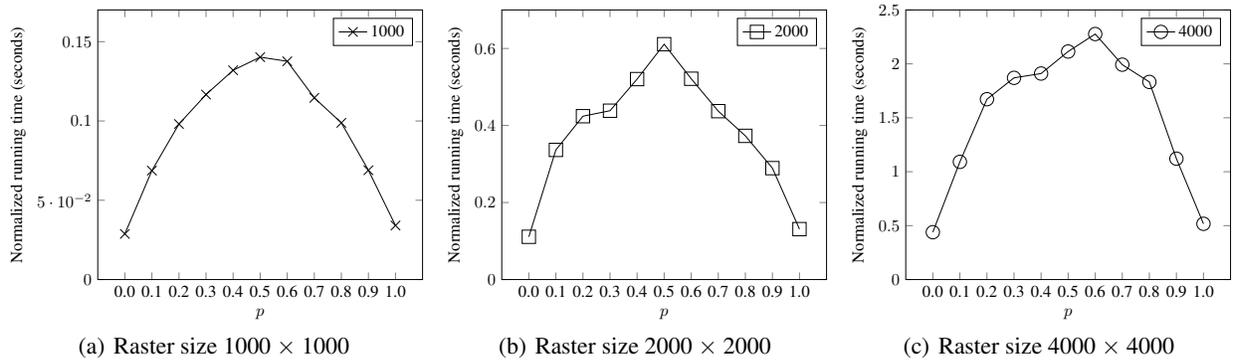
\begin{figure}[!ht]
\centering
\subfigure[Raster size 1000 $\times$ 1000]{\pgfplotstableread{ % data 
Index	0.0	0.1	0.2	0.3	0.4 0.5 0.6 0.7 0.8 0.9 1.0
% 1000 0.013848015	0.034056298	0.044400015	0.04944718	0.055586981	0.055112033	0.054431456	0.046061233	0.040928306	0.028582816	0.014279236
1000 0.02878431259 0.06868757629 0.09807033292 0.11665457096000001 0.13197045797 0.14024285372 0.13762702562 0.11457885199000001 0.09881783929000001 0.06892798912 0.0340783864
}\testdata

\pgfplotstabletranspose[string type,
    colnames from=Index,
    input colnames to=Index
]\testdatanew{\testdata}

\begin{tikzpicture}[scale=0.63]
\begin{axis}[legend pos = north east,
    legend columns=2,
    xlabel={$p$},
    ylabel={Normalized running time (seconds)},
    ymin = 0,
    ymax = 0.17,
    xtick=data,
    xticklabels from table={\testdatanew}{Index},
]
\addplot[mark=x, mark size=4]
table[x expr=\coordindex, y index=1] {\testdatanew};
\legend{1000}
\end{axis}
\end{tikzpicture}}
\subfigure[Raster size 2000 $\times$ 2000]{\pgfplotstableread{ % data 
Index	0.0	0.1	0.2	0.3	0.4 0.5 0.6 0.7 0.8 0.9 1.0
% 2000 0.049018011	0.114110997	0.162259997	0.193434878	0.215874822	0.218345955	0.23040164	0.191858326	0.152921274	0.113867571	0.014279236
2000 0.1110443584 0.33664813860000004 0.42414985847000003 0.43838255098000006 0.5205018332500001 0.6110612413100001 0.5213845531800001 0.4372658299400001 0.37311080973000005 0.28927221917 0.13095006614000002
}\testdata

\pgfplotstabletranspose[string type,
    colnames from=Index,
    input colnames to=Index
]\testdatanew{\testdata}

\begin{tikzpicture}[scale=0.63]
\begin{axis}[legend pos = north east,
    legend columns=2,
    xlabel={$p$},
    ylabel={Normalized running time (seconds)},
    ymin = 0,
    ymax = 0.7,
    xtick=data,
    xticklabels from table={\testdatanew}{Index},
]
\addplot[mark=square, mark size=4]
table[x expr=\coordindex, y index=1] {\testdatanew};
\legend{2000}
\end{axis}
\end{tikzpicture}}
\subfigure[Raster size 4000 $\times$ 4000]{\pgfplotstableread{ % data 
Index	0.0	0.1	0.2	0.3	0.4 0.5 0.6 0.7 0.8 0.9 1.0
% 4000 0.192338864	0.438812999	0.694290652	0.910189194	0.942057948	0.976710899	1.013529768	0.844505799	0.746014666	0.51620624	0.243041517
4000 0.43910372924 1.09140266043 1.67236047665 1.8704450543000002 1.9104259363400002 2.11456818526 2.27661113623 1.99260901652 1.83249983928 1.12086707817 0.5170023890500001 
}\testdata

\pgfplotstabletranspose[string type,
    colnames from=Index,
    input colnames to=Index
]\testdatanew{\testdata}

\begin{tikzpicture}[scale=0.63]
\begin{axis}[legend pos = north east,
    legend columns=2,
    xlabel={$p$},
    ylabel={Normalized running time (seconds)},
    ymin = 0,
    ymax = 2.5,
    xtick=data,
    xticklabels from table={\testdatanew}{Index},
]
\addplot[mark=o, mark size=4]
table[x expr=\coordindex, y index=1] {\testdatanew};
\legend{4000}
\end{axis}
\end{tikzpicture}}
\caption{Running time with different raster and occupied pixels size}
\label{fig:experiment}
\end{figure}  

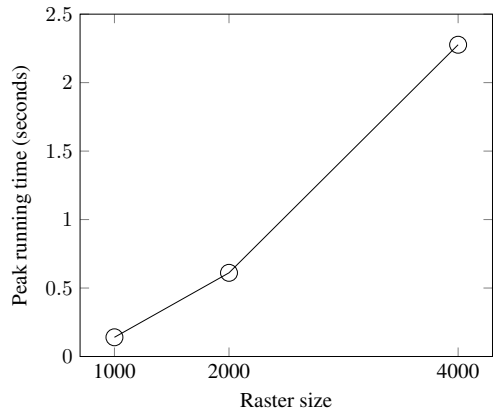
\begin{figure}[!ht]
\centering
\subfigure{\pgfplotstableread{ % data 
Index	1000 2000 4000
% 4000 0.192338864	0.438812999	0.694290652	0.910189194	0.942057948	0.976710899	1.013529768	0.844505799	0.746014666	0.51620624	0.243041517
time 0.14024285372 0.6110612413100001 2.27661113623
}\testdata

\pgfplotstabletranspose[string type,
    colnames from=Index,
    input colnames to=Index
]\testdatanew{\testdata}

\begin{tikzpicture}[scale=0.8]
\begin{axis}[
    xlabel={Raster size},
    ylabel={Peak running time (seconds)},
    ymin = 0,
    ymax = 2.5,
    xtick=data,
    xticklabels from table={\testdatanew}{Index},
]
\addplot[mark=o, mark size=4]
table[x index=0, y index=1] {\testdatanew};
\end{axis}
\end{tikzpicture}}
\caption{Peak running time with different occupied raster size}
\label{fig:experiment_max}
\end{figure}  
We run all experiments on a single machine with Intel Xeon E3-1220 v5 3.00GHz quadcore processor, 64GB RAM, and 2TB HDD on Ubuntu 16.04.2 applied Java 1.8.0\_102. 

In this part, we run some basic experiments to confirm the scalability of the proposed algorithm. We generate random marked rasters of resolutions $1000\times 1000$ up-to $4000\times 4000$ as shown in \autoref{tab:table}. For each raster size, we mark each pixel with an independent Bernoulli distribution with parameter $p$. In other words, we scan over all the pixels and randomly mark each pixel with a probability $p\in[0,1]$. The higher the value of $p$, the more pixels will be marked in the raster. In this experiment, we vary $p$ from 0.0 to 1.0 in increments of 0.1. For each value of $p$, we generate 100 random rasters and compute the average running time.

\autoref{fig:experiment} shows the average running time of the proposed algorithm. In each figure, the running time shows a bitonic, bell-shaped, behavior where it starts very small, peaks around the range $[0.4,0.6]$ and then starts to fall down again. This can be explained by the output-sensitivity of the algorithm. For both very small and very large values of $p$, there are only a very few vertices to be detected since the entropy is low. Thus, the ring formation step of the algorithm finishes very quickly since it will have a few vertices to trace. When the entropy peaks at 0.5, the algorithm takes the longest running time since it will detect the largest number of vertices. This behavior confirms our analysis of output-sensitivity. In reality, when there are real objects to be detected in an image, the entropy will be low and hence the algorithm will run much faster than the peak running time in the figure. The entropy is the highest when the image is purely random which is not expected in real scenarios.

To further evaluate the scalability of the algorithm, \autoref{fig:experiment_max} shows the maximum running time, i.e., at $p=0.5$, as the raster size increases from $1000\times 1000$ to $4000\times 4000$. As expected, the largest running time on the $y$-axis increases linearly with the resolution. For example, when the number of pixels increases from one million to 16 million, the running time increases from 0.14 seconds to 2.28 seconds, i.e., 16 fold.

%it indicates a curve starts to climb up from the beginning and then falls. The time curve reaches the highest point when threads are between 0.4 and 0.6 while taking the least time at 0.0 and 1.0. We notice that fewer pixels are occupied as the thread number reduces. The other observation is that the algorithm detects fewer open vertices when the thread grows. It is expected because 0 indicates no pixel is occupied in the current raster, and 1 means all pixels in raster size are marked. Therefore these two cases reflect the time complexity of the first phase in the proposed approach. It confirms the analysis that the time complexity is bounded by raster width and height. Pixel threads from 0.4 to 0.6 imply more open vertices. Therefore requires more time delineating vertices into rings, although on the same raster. Regardless, desired polygons become less complex if threads are beyond this range. We notice that more vertices are ignored as cases 0 when threads are smaller and skip more cases as 3,5,10,12,15 if threads evolve larger than the range. 

\bibliographystyle{unsrt}  

\bibliography{references}

\begin{thebibliography}{1}

\bibitem{analysis}
Anju Asokan and J~Anitha.
\newblock Machine learning based image processing techniques for satellite
  image analysis -a survey.
\newblock In {\em 2019 International Conference on Machine Learning, Big Data,
  Cloud and Parallel Computing (COMITCon)}, pages 119--124, 2019.

\bibitem{detection}
Gong Cheng and Junwei Han.
\newblock A survey on object detection in optical remote sensing images.
\newblock {\em ISPRS Journal of Photogrammetry and Remote sensing}, 117:11--28,
  2016.

\bibitem{OpenStreetMap}
John~E Vargas~Munoz, Devis Tuia, and Alexandre~X Falc{\~a}o.
\newblock Deploying machine learning to assist digital humanitarians: making
  image annotation in openstreetmap more efficient.
\newblock {\em International Journal of Geographical Information Science},
  35(9):1725--1745, 2021.

\bibitem{polygonization}
Onur Tasar, Emmanuel Maggiori, Pierre Alliez, and Yuliya Tarabalka.
\newblock Polygonization of binary classification maps using mesh approximation
  with right angle regularity.
\newblock In {\em IGARSS 2018-2018 IEEE International Geoscience and Remote
  Sensing Symposium}, pages 6404--6407. IEEE, 2018.

\bibitem{potrace}
Peter Selinger.
\newblock Potrace: a polygon-based tracing algorithm.
\newblock {\em Potrace (online), http://potrace. sourceforge. net/potrace. pdf
  (2009-07-01)}, 2, 2003.

\bibitem{PolyFit}
Edoardo~Alberto Dominici, Nico Schertler, Jonathan Griffin, Shayan Hoshyari,
  Leonid Sigal, and Alla Sheffer.
\newblock Polyfit: Perception-aligned vectorization of raster clip-art via
  intermediate polygonal fitting.
\newblock 39(4), jul 2020.

\bibitem{raptor}
Ahmed Eldawy, Lyuye Niu, David Haynes, and Zhiba Su.
\newblock Large scale analytics of vector+raster big spatial data.
\newblock In {\em Proceedings of the 25th ACM SIGSPATIAL International
  Conference on Advances in Geographic Information Systems}, SIGSPATIAL '17,
  New York, NY, USA, 2017. Association for Computing Machinery.

\end{thebibliography}

% George Kour and Raid Saabne.
% \newblock Real-time segmentation of on-line handwritten arabic script.
% \newblock In {\em Frontiers in Handwriting Recognition (ICFHR), 2014 14th
%   International Conference on}, pages 417--422. IEEE, 2014.

% George Kour and Raid Saabne.
% \newblock Fast classification of handwritten on-line arabic characters.
% \newblock In {\em Soft Computing and Pattern Recognition (SoCPaR), 2014 6th
%   International Conference of}, pages 312--318. IEEE, 2014.

% \bibitem{hadash2018estimate}
% Guy Hadash, Einat Kermany, Boaz Carmeli, Ofer Lavi, George Kour, and Alon
% \newblock Estimate and replace: A novel approach to integrating deep neural
%   networks with existing applications.
% \newblock {\em arXiv preprint arXiv:1804.09028}, 2018.

% \end{thebibliography}

\end{document}